\pdfoutput=1

\documentclass[11pt]{article}

\usepackage[final]{acl}

\usepackage{times}
\usepackage{latexsym}

\usepackage[T1]{fontenc}

\usepackage[utf8]{inputenc}
\usepackage{todonotes}
\usepackage{microtype}

\usepackage{inconsolata}

\usepackage{booktabs}
\usepackage{graphicx}
 \usepackage{amsmath}
 \usepackage{amssymb}
 \usepackage{times}
 \usepackage{amssymb}
 \usepackage{url}
 \usepackage{hyperref}
 \usepackage{amsmath}
 \usepackage{multirow}
 \usepackage{multirow}
 \usepackage{colortbl} 
 \usepackage{float} 
 \usepackage{caption}
 \usepackage{subcaption}
 \usepackage{wrapfig}
\usepackage{xspace} 
\usepackage[ruled,vlined,]{algorithm2e} 
\SetAlFnt{\small}
\newcommand\mydots{\ifmmode\ldots\else\makebox[1em][c]{.\hfil.\hfil.}\thinspace\fi}
\newcommand{\myparagraph}[1]{\noindent \textbf{#1}.}

\usepackage{nicematrix,tikz}

\NiceMatrixOptions
  {
    custom-line = 
     {
       letter = : ,
       command = dashedline , 
       ccommand = cdashedline ,
       tikz = dashed
     }
  }
\title{Learning from Relevant Subgoals in Successful Dialogs using Iterative Training for Task-oriented Dialog Systems}

\author{
 \textbf{Magdalena Kaiser\textsuperscript{1}},
 \textbf{Patrick Ernst\textsuperscript{2}},
 \textbf{Gyuri Szarvas\textsuperscript{2}}
\\
 \textsuperscript{1} Max Planck Institute for Informatics, Saarland Informatics Campus, Germany\\
 \textsuperscript{2} Amazon, Berlin, Germany
\\
mkaiser@mpi-inf.mpg.de, \{peernst, szarvasg\}@amazon.de
}
\newcommand{\suit}{\textsc{Suit}\xspace}
\newcommand{\mars}{\textsc{Mars}\xspace}
\newcommand{\diactod}{\textsc{DiactTOD}\xspace}
\newcommand{\krls}{\textsc{KRLS}\xspace}

\begin{document}

\maketitle

\begin{abstract}
Task-oriented Dialog (ToD) systems have to solve multiple subgoals to accomplish user goals, whereas feedback is often obtained only at the end of the dialog.
In this work, we propose \suit  (= SUbgoal-aware ITerative Training), 
an iterative training approach for improving ToD systems.
We sample dialogs from the model we aim to improve and determine subgoals that contribute to dialog success using distant supervision to obtain high quality training samples. 
We show how this data improves supervised fine-tuning or, alternatively, preference learning results.
Performance improves when applying these steps over several iterations:
\suit reaches new state-of-the-art performance on a popular ToD benchmark.
\end{abstract}

\section{Introduction}
\begin{figure} [t]
    \includegraphics[width=0.92\columnwidth]{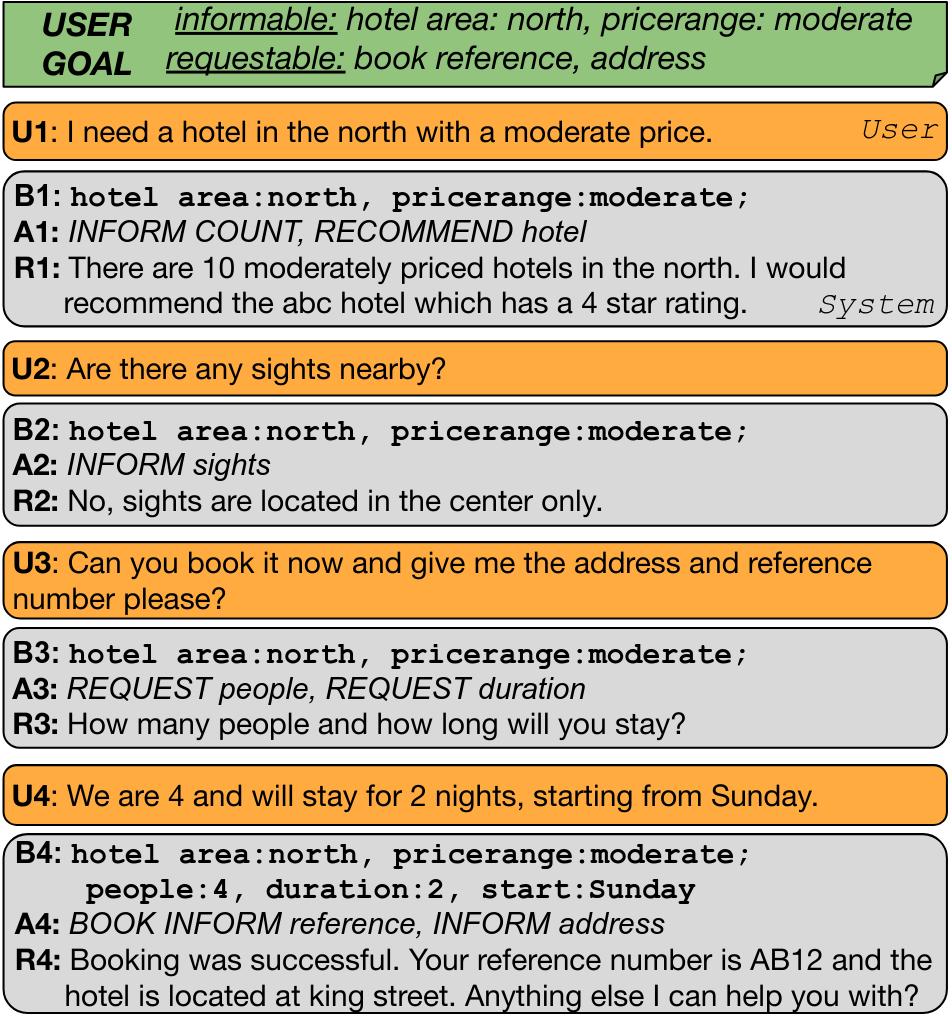}
	\caption{Successful dialog example.}
	\label{fig:runningexample}
\end{figure}

Task-oriented Dialog (ToD) systems nowadays converse with users in natural language conversations and assist them in various tasks, such as booking restaurants, querying weather forecasts and resolving customer service issues.
Fig.~\ref{fig:runningexample} shows a sample conversation for making a hotel reservation, where the user's goal is defined as a set of constraints (\textit{informable slots}, e.g. pricerange) and information needs (\textit{requestable slots}, e.g. hotel address) that should be satisfied at the end of the dialog.
For accomplishing such goals, ToD systems must be able to solve multiple subproblems: (1) dialog state tracking (DST) -- understanding user utterances and keeping track of the conversation by storing relevant information in a structured representation of the dialog progress (belief states $b_i$, e.g. \emph{area} or \emph{price range} in Fig.~\ref{fig:runningexample}), (2) inferring how to react by selecting some dialog actions like database look-ups or requesting more information from the user (actions $a_i$ like \emph{REQUEST number}, \emph{INFORM address} in Fig.~\ref{fig:runningexample}), (3) formulating a natural language response based on the dialog state and actions (like asking about the length of the stay and the number of people in Fig.~\ref{fig:runningexample}, $r_i$).
Specialized approaches focus on solving specific problems, e.g., amongst others, \cite{lee-etal-2021-dialogue}, AG-DST~\cite{tian-etal-2021-amendable}, D3ST~\cite{DBLP:journals/corr/abs-2201-08904} focus on DST, 
LarL~\cite{zhao-etal-2019-rethinking}, TCUP~\cite{Vlastelica_Ernst_Szarvas_2023} concentrate on response generation.
End-to-end (E2E) systems try to leverage complex models to solve all problems jointly. 
Modern ToD systems in that category are based on pre-trained Large Language Models (LLMs) and cast dialog state, action and response generation into sequence prediction problems.
SimpleTOD~\cite{hosseini2020simple} was the first approach which successfully applied this paradigm, by training a causal language model using Supervised Fine-tuning (SFT).
A challenge for ToD systems is the fact that ultimate success with respect to the user's goal is observed at the end of the dialog. 
While Reinforcement Learning (RL) approaches~\cite{zhao-etal-2019-rethinking, lubis-etal-2020-lava, Vlastelica_Ernst_Szarvas_2023} optimize for such sparse rewards, most LLM-based systems neglect these signals and only optimize next turn predictions.  \\
\noindent\textbf{Contributions.} We introduce \suit (= SUbgoal-aware ITerative Training), an E2E ToD system based on LLMs, which contrary to prior work learns from dialog-level success signals. 
Due to the sparseness of these signals, it is unclear which turns, states, actions and responses contribute to the overall success of the dialog. 
For example, the second turn in Fig.~\ref{fig:runningexample} is irrelevant for the success of the dialog. 
The user's goal does not depend on the availability of sights in close vicinity to the hotel, contrary to the respective price range, which will affect the system's success.
We aim to identify these important subparts, which we call \textit{subgoals}, from multiple dialog variants generated by an LLM. 
A naive approach would consider all generations from successful dialogs for SFT, or pair all generations from successful with unsuccessful dialogs for preference learning algorithms, such as RLHF~\cite{NIPS2017_d5e2c0ad} or Direct Preference Optimization (DPO)~\cite{rafailov2024direct}. 
However, these naive approaches cannot distinguish between subgoals that are relevant for the final goal from those that are not.
We employ an iterative distant supervision approach to identify these subgoals that play a major role in dialog success to obtain relevant training samples.
Our contributions are as follows:
\begin{enumerate}
    \item We propose a sampling approach for finding subgoals using distant supervision without relying on external feedback.
    \item We introduce an iterative training procedure for ToD systems.
    \item Our simple but effective approach surpasses state-of-the-art performance on a popular ToD benchmark.
\end{enumerate}
\section{\suit Training Approach}
\begin{figure*} [ht]
\centering
    \includegraphics[width=\textwidth]{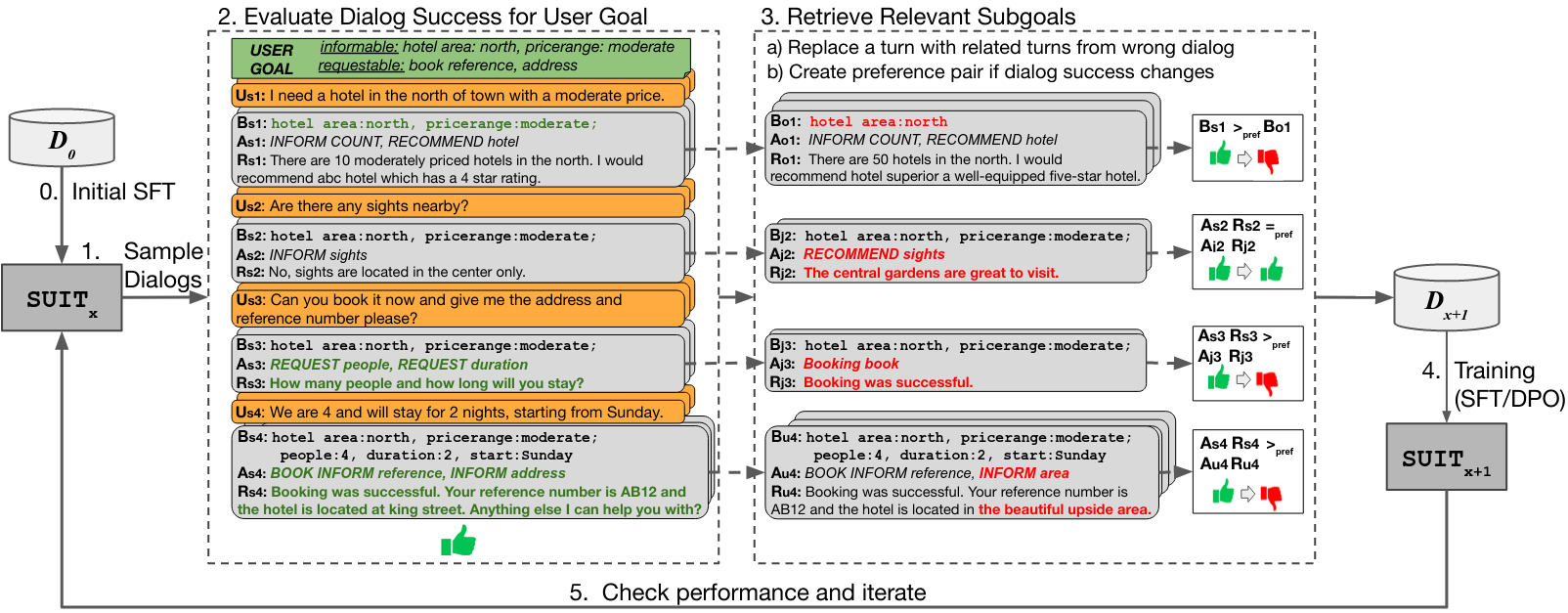}
	\caption{Overview of training procedure in \suit. We sample multiple dialogs for one user goal, where each dialog \textbf{$D_s$} consists of user turns \textbf{$U_{st}$}, and system turns, which are split into dialog states \textbf{$B_{st}$}, system actions \textbf{$A_{st}$} and responses \textbf{$R_{st}$}. We evaluate dialog success at the end of each generated dialog. For every successful dialog \textbf{$D_s$}, we  replace parts of system turns (subgoals) with the respective parts coming from wrong dialogs \textbf{$D_{o,j,u}$}. If the dialog success flips to unsuccessful, we add the successful subgoal as training data.}
	\label{fig:workflow}
\end{figure*}
Fig.~\ref{fig:workflow} provides an overview of the \suit approach.
As input we rely on a ToD dataset $\mathbb{D}$, where each dialog is associated with a user goal. This goal describes the user's information need and a set of constraints, that should be fulfilled at the end of the dialog.
First, an initial LLM is trained on $\mathbb{D}$ with Supervised Fine-tuning (SFT), using turn-level supervision  (Step $0$).
Then, we sample from this model to create dialog variants for each user goal in the training data (Step $1$). 
Next, we determine dialog success for the newly created dialogs (Step $2$).
For each successful dialog, we apply distant supervision to identify subgoals that contribute to the ultimate success of the dialog, by comparing them with generations coming from unsuccessful dialogs (more details in Sec.~\ref{sec:distsupervision}). The subgoals considered relevant comprise further training samples  (Step $3$).
The \suit approach can apply SFT or preference learning based on a dataset which pairs subgoals with negative examples from unsuccessful dialogs (Step $4$).
This procedure is repeatable by sampling from the newly obtained model once again (Step $5$).
Using this effective training paradigm, \suit improves SOTA performance (see Sec.~\ref{sec:experiments}). 
Compared to prior E2E ToD systems, \suit is not based on model customization and does not require feedback from reward models or annotators. The fact, that any off-the-shelf LLM can be plugged into \suit, makes it simple to set up and use in large scale applications.
For preference learning we apply DPO due to its efficiency, low complexity and stable training. 
\subsection{Initial LLM-based ToD Model}\label{sec:initialllm}
For each turn index $t$ in a dialog $D_i$, we are given a dialog context $C_{it} = [U_{i0}, S_{i0}, \ldots, S_{it-1}, U_{it}]$ consisting of the current $U_{it}$, and previous user utterances $U_{ij_{j<t}}$, as well as system turns $S_{ij_{j<t}}$. 
The goal is to train an initial LLM ($\suit_0$) for generating the system turn $S_{it}$, which contains belief states $B_{it}$, system actions $A_{it}$, and a response $R_{it}$.
We split the problem into two prediction tasks:\\
(1) We predict the belief state $B_{it} = \suit_0(C_{it})$;\\
(2) Actions and responses are jointly generated $T_{it} = \suit_0([C_{it}, B_{it}])$, where $T_{it} = [A_{it}, R_{it}]$;\\
We prompt the model twice and introduce special tokens indicating different parts in the generation (see Appendix~\ref{app:modeldetails}).
The model is trained to minimize the negative log-likelihood over the training dataset.
We provide ground truth belief for action and response prediction at training time. 
During inference, \suit generates belief states conditioned on input dialog contexts. Actions and responses are then predicted conditioned on the context and this generated belief. 
\subsection{Subgoal Candidate Generation}\label{sec:distsupervision}
We split each training dialog $D_i$  into all possible dialog contexts $C_{it}$ and
sample a \emph{set} of $k$ states $\mathbb{B}_{it}$, and per state, another $k$ actions $\mathbb{A}_{it}$ and responses $\mathbb{R}_{it}$ from a given $\suit$ model\footnote{We enforce that the samples contain greedy generations}:
\begin{equation*}
{\displaystyle \mathbb{B}_{it} = \bigcup_{B \sim \suit(C_{it})}^{k} B, \mathbb{T}_{it} = \bigcup_{B \in \mathbb{B}_{it}, T \sim \suit(C_{it},B)}^{k} T}
\end{equation*}
Alg.~\ref{algo:suit} shows the precise steps of the \suit training loop.
We obtain $k^2$ candidate dialogs $\mathbb{D}_c$ by replacing the turns $S_{it}$ with samples $S^\prime_{it}$ (1 in Alg.~\ref{algo:suit} and Fig.~\ref{fig:workflow}).
\newcommand\mycommfont[1]{\ttfamily\textcolor{blue}{#1}}
\SetCommentSty{mycommfont}
\SetKwComment{Comment}{\quad}{} 
\begin{algorithm}
	\DontPrintSemicolon
	\SetNoFillComment
	\KwIn{dialogs $\mathbb{D}_x=\mathbb{D}$, $x=0$, user goal partitioning over dialogs $g(\mathbb{D})$, initial model $\suit_x$, $train \in \{DPO, SFT\}$, success criteria $succ$}
	\Repeat{$\suit_{x}$ does not improve\Comment*[h]{(5)}}{
	$\mathbb{D}_{c} \leftarrow \bigcup_{D_i \in \mathbb{D}} \{ 
        [U_{i0}, \mydots, S^\prime_{it}, \mydots, S^\prime_{i|D_i|}] |$
			$S^\prime_{it} = [B_{it}, T_{it}] \sim \suit_x \}$ \Comment*[h]{(1)}\; 
	$\mathbb{D}_{x+1} \leftarrow \emptyset$\;
	\ForEach{dialog set with same goal $\mathbb{D}_{G} \in g(\mathbb{D}_c)$}{
		\ForEach{dialog $D_s \in \mathbb{D}_{G}$: $succ(D_s)$\Comment*[h]{(2)} \ }{
                $\mathbb{D}_{x+1} \leftarrow \mathbb{D}_{x+1} \cup \{(C_{st}, S^\prime_{s\mathbf{t}}) |$  \Comment*[h]{(3)} \; 
                $\exists D_{o} \in \mathbb{D}_{G}, S^\prime_{o\textbf{t}} \in D_{o}: \neg succ(D_{o}) \wedge$ \; $\neg succ([U_{s0}, \mydots, S^\prime_{o\textbf{t}}, \mydots, S^\prime_{s|D_s|}])\}$\;}}
	$\suit_{x+1} \leftarrow train(\suit_{x}, \mathbb{D}_{x+1})$\Comment*[h]{(4)}\;
	$x \leftarrow x + 1$\;
	\caption{\suit Training Approach\label{algo:suit}}}
\end{algorithm}
\subsection{Distant Supervision for Subgoal Detection}\label{sec:subgoaldetection}
We only consider successful dialogs as source of potential training data (2 in Alg.~\ref{algo:suit} and Fig.~\ref{fig:workflow}). 
To determine dialog success, we use the evaluation function from~\cite{nekvinda-dusek-2021-shades} by checking if both INFORM and SUCCESS metrics are fulfilled after the last turn. 
More precisely, a dialog is considered successful if the last offered entity satisfies the user's goal constraints and the system mentioned all \textit{requestable} slots defined in the user's goal in its response. 
In Fig.~\ref{fig:workflow} and Alg.~\ref{algo:suit}, dialog $D_s$ is successful. 
For each successful dialog, we search for unsuccessful dialogs sharing the same user goal (3 in Alg.~\ref{algo:suit} and Fig.~\ref{fig:workflow} ($j,o,u$)). 
If found, we go over the successful dialogs turn-wise and replace 
state $B_{st}$ (and action/response, $A_{st}R_{st}$) with the respective state (action/response) in the unsuccessful one. 
After each replacement, we once again evaluate the modified dialog. 
If the dialog is now unsuccessful, the replaced subgoal was indeed crucial for making it successful. 
If the dialog is still successful, we cannot make any judgement, since the replacement from the unsuccessful dialog might be correct (there can be correct subgoals in unsuccessful dialogs). 
Therefore, we sample replacements from different unsuccessful dialogs to see whether the evaluation changes.
Please note, that we only make \textit{one} replacement at a time, while the other turns of the successful dialog remain unchanged. 
State replacements are done separately, while actions and responses are replaced jointly.
Replacements for a respective turn $t$ come from another dialog (with same user goal) at the same turn level $t$. This makes sense in our setup, since samples share the same ground truth dialog context. Nevertheless, our method is robust to different dialog flows. For high variations in dialog flow, one could additionally apply a similarity based scoring to find the most suitable turn for replacement first.
In~Fig.~\ref{fig:workflow}, replacing the state at turn $1$, as well as replacing action/responses at turn $3$ and $4$ each change the evaluation of  dialog $s$ from successful to unsuccessful, whereas the replacement with $A_{j2}$/$R_{j2}$ results in no change.
Therefore, $B_{s1}$, $A_{s3}$, $R_{s3}$,  $A_{s3}$, $R_{s3}$ are considered as relevant subgoals and will be used for training, while $A_{s2}$ and $R_{s2}$ are \textit{not} used as training data, since no replacement was found that changed the evaluation of the dialog. 
This procedure creates a small, high-quality training set, $\mathbb{D}_{x+1}$ (4 in Alg.~\ref{algo:suit} and Fig.~\ref{fig:workflow}), which is dense in samples that are critical to the final dialog success.
For SFT, we use the selected subgoals, while for DPO, we take the selected subgoals as preferred samples (like $B_{s1}$) and the replacements, which made the dialog change from successful to unsuccessful, as dispreferred samples (like $B_{o1}$).
In summary, \suit's iterative training approach consists of the following steps:\\
(1) Given an LLM $\suit_x$, generate more dialog variations $\mathbb{D}_{c}$ by sampling (see Sec.~\ref{sec:distsupervision});\\
(2) Evaluate $success$ for all dialogs using the evaluation function from \cite{nekvinda-dusek-2021-shades};\\
(3) Identify relevant subgoals by replacing successful ones using distant supervision (see Sec.~\ref{sec:subgoaldetection});\\
(4) Apply SFT or DPO to derive model $\suit_{x+1}$;\\
(5) Assess model performance, repeat or stop; 
\section{Experiments}\label{sec:experiments}
\begin{table*}[ht]
    \centering
        \newcolumntype{G}{>{\columncolor [gray] {0.90}}c}
        \resizebox{0.8\textwidth}{!}{
        \begin{tabular}{l | G  c   G   c   G}
        \toprule
        \textbf{Model} & BLEU & INFORM & SUCCESS & COMBINED& \ \#samples \\ 
         \toprule
        \mars~\cite{sun-etal-2023-mars} & $19.90$ & $88.9$ & $78.0$ & $103.4$ & - \\
        \krls~\cite{yu-etal-2023-krls} & $19.00$ & $89.2$ & $80.3$ & $103.8$ & - \\
       \diactod~\cite{wu-etal-2023-diacttod} & $17.50$ & $89.5$ & $84.2$ & $104.4$ & -\\ \midrule \midrule
        $\suit_0$ (initial SFT)  & $\boldsymbol{19.94}$ & $80.4$ & $72.5$ & $96.39$ & - \\ \midrule
        + all subgoals (SFT) & $19.50$ & $87.0$ & $79.4$ & $102.70$ & $31586$ \\
        + all subgoals (DPO) & $17.79$ & $86.9$ & $80.6$ & $101.54$ & $31586$ \\ \midrule 
          $\suit_1$ (SFT)  & $17.75$ & $89.8$ & $84.0$ & $104.65$ &  $4838$\\
          $\suit_1$ (DPO) & $17.44$ & $88.5$ & $82.7$ & $103.04$ &   $4838$ \\ \midrule
          $\suit_2$ (SFT-SFT) & $15.11$ & $89.7$ & $85.9$ & $102.91$ & $2493$  \\
          $\suit_2$ (SFT-DPO) & $17.17$ & $89.5$ & $84.4$ & $104.12$ &  $2493$  \\
          $\suit_2$ (DPO-SFT) & $16.47$ & $\boldsymbol{90.0}$ & $\boldsymbol{87.1}$ & $\boldsymbol{105.02}$ &  $2166$ \\
          $\suit_2$ (DPO-DPO)   & $16.92$ & $88.8$ & $84.4$ & $103.52$ &  $2166$  \\
        \bottomrule
        \end{tabular}
        }
       \caption{\suit~results compared to other SOTA systems on MultiWOZ 2.2.}
        \label{tab:main}
        \end{table*}
\myparagraph{Dataset} 
We use MultiWOZ 2.2~\cite{zang2020multiwoz}, which is a popular ToD benchmark. It contains 10k human-human dialogs over 7 domains.\\
\myparagraph{Metrics} 
We follow the standardized evaluation from \cite{nekvinda-dusek-2021-shades} to allow for a better comparability. 
A delexicalized BLEU score measures response coherence, while INFORM and SUCCESS rates express how much a user's goal is fulfilled at the end of the dialog. 
It is common to assess the overall performance with a COMBINED score $=BLEU+\frac{INFORM+SUCCESS}{2}$.\\
\textbf{Model.} We use an encoder-decoder Flan-T5 large model, which is trained for 1 epoch per iteration.
We verbalize states and actions to be more suitable for generative models. 
Examples for this verbalization and hyperparameters can be found in App.~\ref{app:modeldetails}.\\
\myparagraph{Baselines}
We compare with state-of-the-art E2E systems from the MultiWOZ leaderboard.
\mars~\cite{sun-etal-2023-mars} uses a contrastive loss to differentiate dialog contexts with the same states from dissimilar ones. 
\krls~\cite{yu-etal-2023-krls} applies RL with a special reward, sensitive to important keywords.
\diactod~\cite{wu-etal-2023-diacttod} uses an action encoder to perform nearest neighbor search over latent representations of ground truth actions.

\subsection{Main Results}
Tab.~\ref{tab:main} shows \suit~models, trained for up to two iterations with SFT/DPO.  
It compares their performances with competitors and baselines, including models trained for one round with all successful dialogs instead of using relevant subgoals only.

\myparagraph{\suit~improves SOTA performance}
Our initial model reaches the highest BLEU score ($19.94$). 
While further iterations decrease this metric slightly, they reach the highest INFORM ($90.0$), SUCCESS ($87.1$), and COMBINED ($105.02$) scores and surpass all E2E competitors on the MultiWOZ leaderboard.
The decreasing BLEU score is unavoidable. Due to sampling, the generations may deviate from the fixed reference responses. The responses, however, are still accurate and fluent.
Both training paradigms (DPO/SFT) show improvements. In fact, best results are reached with a combination (DPO-SFT).

\myparagraph{\suit~identifies the most helpful training data}
Training with all successful dialogs, even though comprising a much larger set ($31586$ vs. $4838$ and $2493$/$2166$), performs worse than training with relevant subgoals only.
This shows the quality degradation of training data, when not selecting relevant subgoals carefully and confirms the benefits of \suit's approach (Sec.~\ref{sec:distsupervision}).

\myparagraph{Sampling new dialogs greatly improves the initial model $\suit_0$}
The INFORM metric improves from $80.4$ to $89.8$ and SUCCESS from $72.5$ to $84.0$ after training for one iteration.
A second iteration further improves SUCCESS from $84.0$ up to $87.1$.
After the second iteration, the COMBINED score is not increasing any further, thus we stop.

\subsection{Domain-wise Results}
Tab.~\ref{tab:domainresults} shows the domain-wise results for INFORM and SUCCESS  of our best \suit models (per iteration).
\suit performs well on all domains.  
While results for the train domain are already quite high for $\suit_0$, results of the other domains are improved a lot over iterations. Especially, there are large jumps for the attraction domain (INFORM from $84.3$ to $97$ and SUCCESS from $68.9$ to $89.4$).

\begin{table*}[ht]
   \centering
       \newcolumntype{G}{>{\columncolor [gray] {0.90}}c}
          \resizebox{0.8\textwidth}{!}{
       \begin{tabular}{l | G c G c  G  |   G  c  G  c G}
       \toprule
       & \multicolumn{5}{c |}{INFORM} & \multicolumn{5}{c}{SUCCESS} \\  \toprule
         \textbf{Model} $\downarrow$ / \textbf{Domain} $\rightarrow$ &  train & attr. & rest. & taxi & hotel  & train & attr. & rest. & taxi & hotel  \\
        \toprule
     $\suit_0$ & $94.5$ & $84.3$  & $88.3$ & $100.0$  & $83.0$  & $78.8$ & $68.9$ & $75.3$ & $72.8$ & $74.9$ \\
        $\suit_1$ (SFT) & $92.1$  & $97.5$ & $96.3$ & $100.0$ & $89.8$ & $86.7$ & $85.4$ & $87.2$ & $86.7$  &  $83.5$ \\
        $\suit_2$ (DPO-SFT) & $92.3$  & $97.2$& $96.1$ & $100.0$  & $90.4$ & $86.7$ & $89.4$ & $89.2$  &  $89.7$ & $84.0$ \\
     \bottomrule
       \end{tabular}
       }
      \caption{Domain-wise results for INFORM and SUCCESS metrics of best \suit model per iteration (\textit{attr.} is short for \textit{attraction} and \textit{rest.} for \textit{restaurant} domain). }
       \label{tab:domainresults}
\end{table*}
\section{Related Work}
\textbf{End-to-end ToD Systems.} 
SimpleTOD~\cite{hosseini2020simple} optimizes all sub-tasks jointly using causal language modeling.
Prior work 
uses specialized losses (MTTOD~\cite{lee-2021-improving-end}, Mars~\cite{sun-etal-2023-mars}) or focus on special encoders (DiactTOD~\cite{wu-etal-2023-diacttod}) or learn adapters for the individual tasks (TOATOD~\cite{bang2023task}).
KRLS~\cite{yu-etal-2023-krls}, CASPI~\cite{ramachandran2022caspi}, CALM~\cite{snell2022context} and RewardNet~\cite{feng2023fantastic}  apply RL with special reward functions.
GALAXY~\cite{he2022galaxy} applies semi-supervised learning and 
in~\cite{steindl2024counterfactual} a data augmentation approach by mixing existing dialogs is proposed, whereas our sampling-based approach additionally enriches lexical variety and carefully selects the most helpful training data using distant supervision.
\\
\textbf{Preference Learning.}
\citet{10.5555/3495724.3495977} use RLHF for fine-tuning.
\citet{kaufmann2024survey} provide an overview of work applying RLHF.
DPO~\cite{rafailov2024direct}, PRO~\cite{Song_Yu_Li_Yu_Huang_Li_Wang_2024}, IPO~\cite{pmlr-v238-gheshlaghi-azar24a}, and RRHF~\cite{NEURIPS2023_23e6f78b} optimize for preferences with supervised learning.
\citet{guo2024direct} study these methods in online setups using LLM annotators.
\citet{xu2023some} adopt Cringe Loss~\cite{adolphs-etal-2023-cringe} to preference data. 
Contrary to our work, these approaches require external feedback.	

\section{Conclusion}
\suit is an iterative training approach for ToD systems, which couples 
sampling to derive new dialogs, with distant supervision to determine subgoals that impact the final dialog success.
This coupling enables \suit models to improve INFORM and SUCCESS metrics and advance the SOTA.

	
\section{Limitations}
One limitation of the current approach is the fact that we rely on evaluating dialog success based on ground truth user goals. 
We do not consider generating new goals, for example by simulating users. 
By only performing turn-wise replacements, the newly generated training samples will closely follow the flow of the ground truth dialogs.
However, for some subgoals order does not matter, e.g., in which order to ask for user preferences is most of the time not crucial for success, and generating them in arbitrary order may increase the diversity of the generated dialogs.
The experiments in this paper focus on MultiWOZ, since it is the most common dataset of task-oriented human-human conversations. 
Due to its adoption and range over multiple domains, MultiWOZ provides a general benchmark for ToD systems. 
However, transferring the learning and methods to a different dataset could further strengthen the generalizability aspect of this work.
We consider the aforementioned limitations for future work.
\section{Ethical Considerations}
There are no negative ethical and societal concerns arising from
this work. The used data is provided by~\cite{zang2020multiwoz} and no further human intervention was required.
We train models based on FlanT5-large ($783M$ parameters), which can be considered as  lightweight compared to much larger GPT/LAMA models, keeping the environmental impact comparatively small.

\bibliography{suitbib}
\appendix

\section{Appendix}
\label{sec:appendix}

\subsection{Experimental Details}\label{app:modeldetails}
\begin{figure*} [ht!]
    \includegraphics[width=0.9\textwidth]{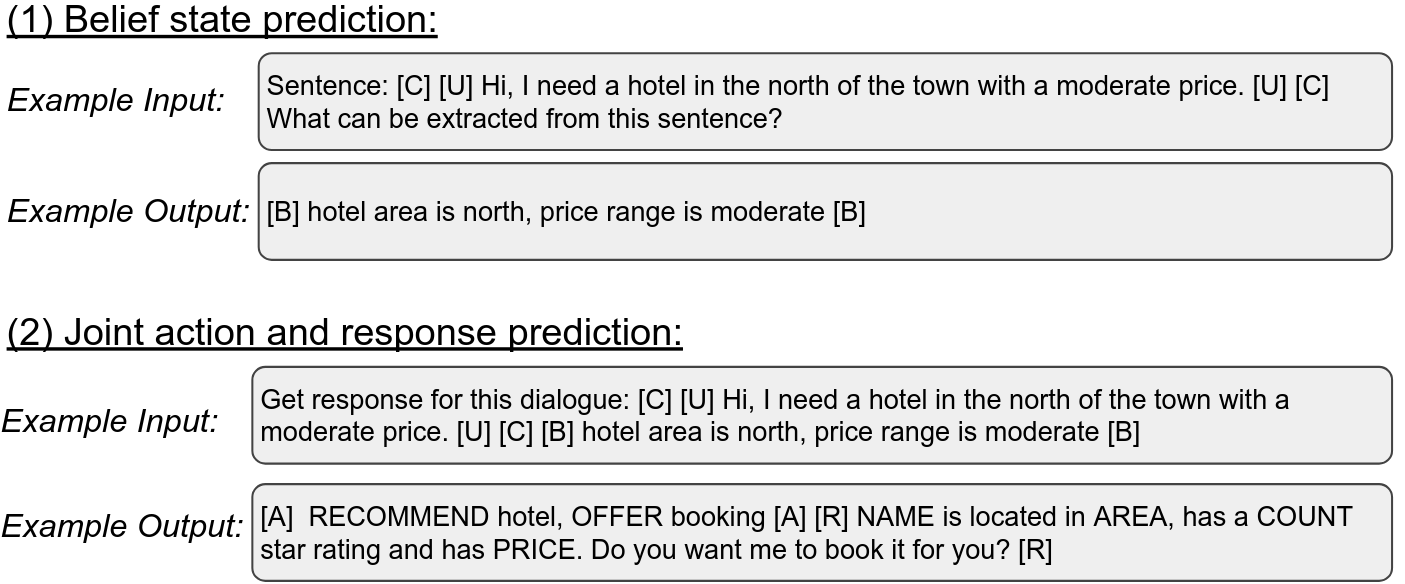}
 \caption{Example for input/ouput representation in \suit.}
	\label{fig:prompts}
\end{figure*}

We use a Flan-T5 large model from Hugging Face\footnote{\url{https://huggingface.co/google/flan-t5-large}} in our experiments.\\
\noindent \textbf{Input/Output Representations.}
As described in Sec.~\ref{sec:initialllm}, we split the generation into two separate prediction tasks:\\
(1) Predicting the belief state: $B_{it} = \suit_0(C_{it})$\\
(2) Jointly predicting actions and responses: $[A_{it}, R_{it}] = \suit_0([C_{it}, B_{it}])$ \\
Fig.~\ref{fig:prompts} shows an example for these predictions. Special tokens $[C], [U], [R], [B], [A]$ are used to indicate dialog context, user utterances, system responses, belief states and actions accordingly. 

\begin{table*}[ht!]
   \centering
       \newcolumntype{G}{>{\columncolor [gray] {0.90}}c}
      \begin{tabular}{l | c | G  c |  G c  G c  G  c  }
      & \# Goals & \multicolumn{2}{c}{\# Dialogs} & \multicolumn{6}{c}{\# Successful Dialogs per Goal} \\ 
         sampled from & & \# suc. & \# unsuc. & $0$ & $1$ & $2$ & $3$ & $4$ & $5$\\
      \toprule
      $\suit_0$ & $4218$ & $7720$  & $13370$ & $1510$  & $590$  & $599$ & $593$ & $477$ & $449$ \\
      $\suit_1$ (SFT) & $4218$  & $11983$ & $9107$ & $1212$ & $212$ & $248$ & $407$ & $641$  &  $1498$ \\
         $\suit_1$ (DPO) & $4218$  & $11831$& $9259$ & $1278$  & $242$ & $236$ & $356$ & $481$  &  $1625$ \\
   \bottomrule
      \end{tabular}
      \caption{Sampling statistics for \suit models.}
      \label{tab:samplestats}
\end{table*}

\noindent \textbf{Hyperparameters.}
After initial model training ($\suit_0$), \suit~models are trained for up to two rounds of SFT/DPO. As stopping criteria we use the COMBINED score. 
For efficiency, we sample half of the user goals in the training data per iteration for creating new dialogs. 
We use $k=2$ for sampling these new dialogs and additionally take the greedy generation (resulting in $k^2 +1$ different dialogs). 

For supervised learning, the batch size was set to $2$ and learning rate to $5e-5$. The maximal input length of $512$ was used, the maximal target length was $256$ and for generation, beam search with a beam size of $5$ was used. We use default parameters when sampling from the model (top-$k$ was set to 0).

For DPO training, we set hyperparameters as follows: $\beta = 0.1$, batch size  $=2$, learning rate  $=1e-6$, warmup steps  $=150$, maximal input length $=512$, and target length $=256$.

\subsection{Data Statistics}\label{app:datastats}
We use the MultiWoZ version 2.2\footnote{\url{https://github.com/budzianowski/multiwoz/tree/master/data/MultiWOZ_2.2}} and for pre-processing and evaluation we follow \url{https://github.com/Tomiinek/MultiWOZ_Evaluation/tree/master}.
Tab.~\ref{tab:multiwoz} shows some statistics about this dataset.
\begin{table}[htb]
   \centering
       \newcolumntype{G}{>{\columncolor [gray] {0.90}}c}
      \resizebox{\columnwidth}{!}{
       \begin{tabular}{l | c| c | c }
        & \# Goals & \#Turns & avg. \#Turns/Dialog \\ 
        \toprule
      train set & $8437$ & $56776$ & $6.70$  \\
     dev set & $1000$ & $7374$ & $7.37$  \\
      test set & $1000$ & $7372$ & $7.37$ \\ \bottomrule
       \end{tabular}
       }
      \caption{Data statistics for MultiWOZ 2.2.}
       \label{tab:multiwoz}
\end{table}

Tab.~\ref{tab:samplestats} shows some statistics on sampling new dialogs in \suit.
For each iteration, half of the user goals ($4218$) in the train set are sampled.
Then, $5$ new dialogs are created for each sampled user goal, resulting in $21090$ newly generated dialogs in total (see amount of dialogs evaluated as succ/unsucc in the table). The right side in Tab.~\ref{tab:samplestats} shows the number of user goals for which there were $0/5$ successful dialogs, $1/5$ successful dialog, ..., up to $5$ out of $5$ successful dialogs. 
Entries in $1$-$4$, which contain at least one successful and one unsuccessful dialog, are the candidates used in \suit, since our approach requires comparison between successful and unsuccessful dialogs for the same user goal.

\begin{table}[ht]
   \centering
       \newcolumntype{G}{>{\columncolor [gray] {0.90}}c}
      \resizebox{\columnwidth}{!}{
      \begin{tabular}{l | c |  G   G}
      & \# Goals & \multicolumn{2}{G}{\# Subgoal Samples} \\ 
      sampled from   & & states & actions/responses \\
      \toprule
      $\suit_0$ & $2238$  & $836$ & $4002$\\
      $\suit_1$ (SFT) & $1461$  & $617$ & $1876$ \\
      $\suit_1$ (DPO) & $1278$ & $746$ & $1420$   \\
   \bottomrule
      \end{tabular}
      }
      \caption{Subgoal train data statistics for \suit models.}
      \label{tab:subgoaldata}
   \end{table}

Tab.~\ref{tab:subgoaldata} shows statistics about the new subgoal-based training data obtained by our approach. We show the number of different user goals present in the new train data as well as the number of subgoals that were considered relevant, split into subgoals representing states and subgoals representing action/response generations.
Additionally, we found that the most relevant turns are turn $2$-$5$ and the top-$5$ relevant dialog acts are: \textit{inform NAME}, \textit{book REFERENCE}, \textit{inform PRICE}, \textit{inform AREA}, \textit{inform PHONE}.



\subsection{Example Dialogs}
\begin{table*} [ht] 
	\centering
  \begin{NiceTabular}{p{14cm}}
\toprule
       
     \textit{Domain: Hotel} \\ \textbf{Context:}  \textbf{User:} Hello! Can you tell me about places to stay in the north area of town? I'll be on a business trip, so I do need free wifi. \textbf{System:} I have several options that meet your needs. I would recommend the Avalon, unless you need free parking. \\  \textbf{User:} I do not need parking, is the Avalon moderately priced? \\ \dashedline
           \textbf{Positive Action:} \textit{ booking hotel inform NAME; inform PRICE; }  \\
               \textbf{Negative Action:} \textit{booking hotel inform PRICE; inform AREA; inform COUNT;}  \\
            \textbf{Positive Response:} \textit{NAME is \textcolor{green}{PRICE}. would you like me to book it for you? }  \\
               \textbf{Negative Response:} \textit{very good! it is in the AREA and has COUNT stars. would you like me to book it for you?} \textcolor{red}{missing PRICE} \\ \midrule \midrule
               
        \textit{Domain: Attraction} \\ \textbf{Context:} \textbf{User:} Hi, I am planning my Cambridge trip and could use some help with a particular attraction. \textbf{System: } I sure can, what is the name of the attraction you are seeking information about? \\  \textbf{User: } I'm looking for tenpin I want the address and entrance fee. \\ \dashedline
           \textbf{Positive Action:} \textit{attraction inform ADDRESS; inform PRICE; inform NAME; inform POST; general  }  \\
               \textbf{Negative Action:} \textit{attraction inform AREA; inform PRICE; inform NAME; general}  \\
            \textbf{Positive Response:} NAME is a nearby attraction, admission is \textcolor{green}{PRICE}, and they are located at \textcolor{green}{ADDRESS}, postcode POST. is there anything else?  \\
               \textbf{Negative Response: } NAME is in AREA and it's \textcolor{green}{PRICE} to enter. can i help you with anything else? \textcolor{red}{missing ADDRESS}
                  \\   \bottomrule
	\end{NiceTabular}  
	\caption{Dialog examples with positive/negative subgoals.}
	\label{tab:dialogexamples2}
\end{table*}
\begin{table*} [t] 
	\centering
		\begin{NiceTabular}{p{14cm}}
  \toprule
   

              \textit{Domain: Train} \\
              \textbf{Context:}  ... \textbf{User:} Okay, sounds good. Also I need to get a train from Liverpool street to Cambridge. \textbf{System:} What day would you like to make this trip? Do you have any other specifications? \\  \textbf{User: } The train should leave after 12:45 and should leave on friday. for 2 people please book. 
          \\ \dashedline
            \textbf{Positive State:} \texttt{train \textcolor{green}{departure: london liverpool street}; \textcolor{green}{destination: cambridge}; }  \\
               \textbf{Negative State:} \texttt{train \textcolor{red}{departure: cambridge};  \textcolor{red}{destination: london liverpool street}; } \textcolor{red}{swapped departure and destination} \\ \midrule \midrule
                                  \textit{Domain: Restaurant} \\ \textbf{Context:} [] \\  \textbf{User: } I'm looking for a restaurant with mediterranean food. \\ \dashedline
           \textbf{Positive Action:} \textit{booking restaurant inform AREA; inform COUNT; inform FOOD; inform NAME; inform PRICE; }  \\
               \textbf{Negative Action: } \textit{restaurant inform COUNT;}  \\
            \textbf{Positive Response:} i have COUNT PRICE FOOD restaurants in the AREA. NAME and NAME. would you like me to book one for you?  \\
               \textbf{Negative Response:} there are COUNT. for booking do you have a preferred information is there a specific time of the day or time of day you would like to arrive by? \textcolor{red}{noisy/less concrete} \\ \midrule \midrule           
      
        \textit{Domain: Taxi} \\ \textbf{Context:}  \textbf{User:} I am traveling to Cambridge and excited about seeing location attractions. Could you help me find a place to go, like a college? \textbf{System:} Certainly. I have many available. There is corpus christi college, fore example ...
        \textbf{User:} I'm also looking for a hotel in the centre. ... \textbf{System:} Your booking at university arms hotel was successful with a reference number of S1HTVV32. Is there anything else I can do for you today? \\
   \textbf{User: } I need a taxi please between the 2 places. I want to leave the attraction by 2:30.
        \\ \dashedline
        \textbf{Positive State:} \texttt{\textcolor{green}{taxi departure: corpus christi; destination: university arms hotel; leave is 02:30};  hotel area: centre; bookday: tuesday; bookstay: 1; name: university arms hotel; attraction type: college;
        } \\
        \textbf{Negative State:} \texttt{hotel area: centre; bookday: tuesday; bookstay: 1; name: university arms hotel; stars: 4; attraction type: college;
       } \\ 
        \textbf{Postive Action:} \textit{ taxi inform PHONE; inform TYPE;} \\
         \textbf{Negative Action:} \textit{taxi request PLACE;} \\
        \textbf{Postive Response:} booking completed! booked car type: TYPE contact number: PHONE\\
        \textbf{Negative Response:}  sure, i can provide you a car if you like. where are you departing from? \textcolor{red}{info from context not considered}\\  \bottomrule
	\end{NiceTabular}  
	\caption{Dialog examples with positive/negative subgoals.}
	\label{tab:dialogexamples1}
\end{table*}
Tab.~\ref{tab:dialogexamples2} and \ref{tab:dialogexamples1} show excerpts from sample dialogs, where relevant subgoals were detected, along with the negative example used in DPO.
Highlighted in \textcolor{green}{green} one can see the relevant information from a subgoal that contributes in reaching dialog success, whereas in \textcolor{red}{red} the information from the unsuccessful dialog is shown, that changed the evaluation of the successful dialog when used as a replacement.

\end{document}